\documentclass{article}

\usepackage{arxiv}

\usepackage[utf8]{inputenc} 
\usepackage[T1]{fontenc}    
\usepackage{hyperref}       
\usepackage{url}            
\usepackage{booktabs}       
\usepackage{amsfonts}       
\usepackage{nicefrac}       
\usepackage{microtype}      
\usepackage{lipsum}		
\usepackage{graphicx}
\usepackage{natbib}
\usepackage{doi}
\usepackage{amsmath}
\usepackage{array}

\usepackage{fixltx2e}

\usepackage{stfloats}

\title{Neuromuscular Modeling for Locomotion with Wearable Assistive Robots - A primer}

\author{Mohamed Irfan Refai$^{1,*}$, Huawei Wang$^1$, Antonio Gogeascoechea$^1$, Rafael Ornelas Kobayashi$^1$, \\
\textbf{Lucas A. Gaudio$^1$, Federica Damonte$^1$, Guillaume Durandau$^{2,3}$,} \\
\textbf{Herman van der Kooij$^1$, Utku S. Yavuz$^4$, Massimo Sartori$^1$} \\ 
        Email: \texttt{$^{*}\{$m.i.mohamedrefai$\}$}@utwente.nl \\
$^{1}$ Department of Biomechanical Engineering, University of Twente, The Netherlands, \\
$^{2}$ Department of Mechanical Engineering, McGill University, Canada, $^{3}$ CRIR - Jewish Rehabilitation Hospital, \\
$^{4}$ Biomedical Signals and Systems, University of Twente, The Netherlands}

\begin{document}
\maketitle

\begin{abstract}
Wearable assistive robots (WR) for the lower extremity are extensively documented in literature. Various interfaces have been designed to control these devices during gait and balance activities. However, achieving seamless and intuitive control requires accurate modeling of the human neuromusculoskeletal (NMSK) system. Such modeling enables WR to anticipate user intentions and determine the necessary joint assistance. Despite the existence of controllers interfacing with the NMSK system, robust and generalizable techniques across different tasks remain scarce. Designing these novel controllers necessitates the combined expertise of neurophysiologists, who understand the physiology of movement initiation and generation, and biomechatronic engineers, who design and control devices that assist movement. This paper aims to bridge the gaps between these fields by presenting a primer on key concepts and the current state of the science in each area. We present three main sections: the neuromechanics of locomotion, neuromechanical models of movement, and existing neuromechanical controllers used in WR. Through these sections, we provide a comprehensive overview of seminal studies in the field, facilitating collaboration between neurophysiologists and biomechatronic engineers for future advances in wearable robotics for locomotion.\\
\end{abstract}



\section{Introduction}
Wearable assistive robots (WR) for the human lower extremity, such as exoskeletons and prostheses, enhance or replace mobility of an impaired limb, providing crucial assistance for individuals with movement disabilities \citep{Awad2020, eilenberg2010control, Kim2019, Lee2017, Mooney2014, Quintero2012, Seo2016, Walsh2011, Wang2015}. For effective and synchronized assistance during movement, WRs require information about the intended human action. 
Although several approaches allow WRs to interface with human intentions \citep{Lobo-prat2014, Tucker2015, Young2017, Fleming2021, Masengo2023, DeMiguel-Fernandez2023}, most rely on tracking pre-defined trajectories \citep{ DeMiguel-Fernandez2023, van2014combined}, or torques \citep{zhang2017human} or damping \citep{johansson2005clinical} profiles. These trajectories reflect the movement generated by the musculoskeletal system. Controllers for WRs that track predefined trajectories monitor the human for triggers to predict subsequent motions or trajectories.

However, these approaches are limited as they are task-dependent and rely on resulting human movement rather than prior intention. WRs, targeting the lower extremity, must address both mobility and balance, which is challenging \citep{roffman2014predictors, Young2017, pinto2020performance}. Human intentions, encoded as neural signals, move the skeletal system via the nervous system and muscles. For intuitive interfacing, WRs should access these information systems, integrating this knowledge into a digital twin of the musculoskeletal system for a task-agnostic model of human motion \citep{Fleming2021}. Since neural processes occur milliseconds before movement, this allows WRs to synchronize better with human motion. Measuring and modeling neural signals and subsequent movements is essential for effective WR interfacing, particularly during reduced mobility due to musculoskeletal disorders or amputation, enhancing or replacing the individual's mobility.

WR controllers inspired by human neurophysiology involving sensory and local circuitries \citep{DAvella2010, Geyer2010, sartori2013musculoskeletal} have shown to be generalisable for locomotion in virtual avatar simulations \citep{Song2015, Song2012}. In practice, neural intention for movement could be extracted from signals such as electromyograms (EMG) \citep{Durandau2018, Fleming2021} and used to drive virtual muscle-tendon models that drive WRs \citep{Durandau2022, durandau2019voluntary}. The spatio-temporal resolution of neural intention can be improved via advanced techniques that record the motor neuron pool \citep{negro2016multi} and subsequently the muscle fiber kinematics \textit{in vivo} \citep{dick2023advances}. Nonetheless, such control systems \citep{Geyer2010, sartori2013musculoskeletal} utilize simple models of the musculoskeletal systems, which limits their application during balance, perturbation, or physical interaction with wearable robots. Moreover, we need to establish a causal relationship between the neural decoding of musculoskeletal forces and the influence of musculoskeletal disorders or amputation. This allows us to target rehabilitation and improve mobility in individuals with movement impairment. 

Gathering the know-how to design such precise and intuitive WRs requires bridging the neurophysiologists and biomechatronists that focus on lower extremity mobility and robotics. Nonetheless, there are review papers that address neural control of locomotion \citep{Cote2018, duchateau2016neural}, or control strategies for lower-limb robots \citep{Lobo-prat2014, Fleming2021, Tucker2015, Young2017, Baud2021}. However, these papers target either the neurophysiologists or biomechatronists more than the other. Thus, there is significant value for a primer that summarizes key concepts for both the neurophysiologists and biomechatronists, but also for neuroscientists, and roboticists that work with the lower extremity. This could inspire the next generation of scientists and researchers to develop new bio-inspired control algorithms for lower extremity WRs. 

Therefore, this article aims to provide fundamental concepts across neurophysiology and biomechatronics that are needed for design of intuitive WRs for the lower extremity. The current article is divided into three parts. The first, part, on the neuromechanics of locomotion, we offer a didactic introduction of the current neurophysiology knowledge related to human locomotion control. The second part, on neuromechanical models describing locomotion, summarizes the mathematical models that describe neuromechanics of locomotion. Finally, the third part, on neuromechanical controls for WRs, describes the state-of-the-art appraoches that utilize the knowledge of the NMSK system for controlling WRs. At the end of the review, we present an outlook for the future.

\section{Neuromechanics of locomotion}

\begin{figure*}
    \centering
    \includegraphics[width=0.7\textwidth]{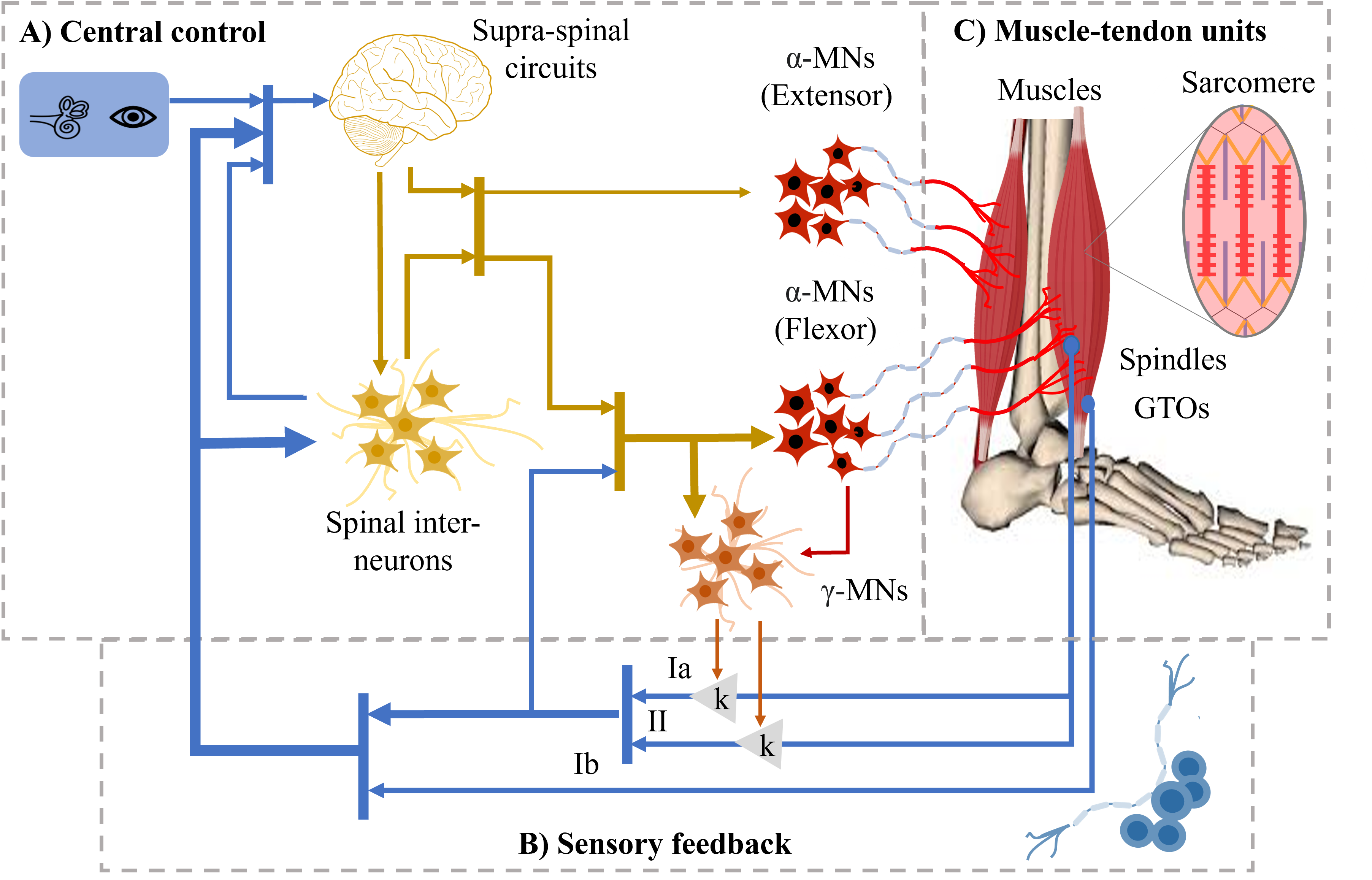}
    \caption{High-level representation of the neural control for human locomotion. A) The central control (\textit{i.e.,} supra-spinal circuits, inter-neurons, etc.) integrates neural information from sensory systems (\textit{e.g.,} vestibular, visual, propioceptive, etc.) and generates a motor command. This is transmitted to pools of $\alpha$-motoneurons, which transform the final common signal and send it to the mechanical actuator (\textit{i.e.,} the muscle-tendon units). B) Sensory feedback from the propioceptive system is provided by Golgi tendon organs (GTOs) and muscle spindles. This is transmitted to the central control via Ia, Ib, and II afferent fibres. Here, only fibres from flexor muscle are depicted for clarity. The sensitivity of the Ia and II fibers (K) is regulated by $\gamma$-motoneurons (\textit{i.e.,} fusimotor system). C) Mechanical actuation results from the contraction of flexor and extensor muscles. These contract in response to the motor code send by $\alpha$-motoneurons. Thicker lines represent cumulation of the thinner pathways.     }
    \label{Fig1_updated}
\end{figure*}

Human locomotion is the result of intricate interactions between the nervous, skeletal, and muscular systems. From a control theory perspective, locomotion can be described as a closed-loop system (Fig. \ref{Fig1_updated}), where the controller (i.e, spinal and supra-spinal neurons) integrates sensory feedback information (i.e., proprioceptive, visual, vestibular and cutaneous sensors) to modulate mechanical actuation (i.e., muscle activation and resulting force). At the intersection between the controller and the actuator, the alpha-motoneurons ($\alpha$-MNs) play a key role in the generation and transmission of the control commands. Thus, to develop personalized wearable technologies, it is of utmost importance to understand the physiological interactions between these systems. The central control, sensory feedback, and muscle-tendon units are described in detail below. 

\subsection{Central Control}

\label{sec:NeuralControl}

Locomotion involves the rhythmical activation of skeletal muscles, modulation of posture, balance, and adaptation to the environment. These processes are controlled by neuronal circuits located within the spinal cord and supra-spinal structures (\textit{e.g.}, brainstem, motor cortex, etc.). 

At the level of the spinal cord, circuits of interneurons, commonly represented as central pattern generators (CPGs), play a key role in the coordination of the rhythmical activity of lower-limb muscles \citep{lacquaniti2012patterned}. For instance, during the stance phase, the activation of the quadriceps muscle (\textit{i.e.}, knee extension) triggers the alternating activity of Soleus and Tibialis anterior (\textit{i.e.,}, ankle plantar- and dorsi-flexion, respectively) \citep{lamy2008}. The control of these processes has been suggested to take place almost entirely at the spinal level, purely driven by CPGs and muscle-level sensory feedback. This was shown from experiments on a decerebrated cat walking on a treadmill, where spinal circuits were able to generate stable gait patterns in absence of any brain input \citep{Capogrosso2016, Whelan1996}. Moreover, human subjects exhibiting complete spinal cord injury (\textit{i.e.,} spinal neurons deprived of any brain input) were shown to produce alternating flexion-extension activity patterns when triggering a flexor reflex \citep{BUSSEL1988}.

The neuronal circuits in the spinal cord controlling these processes include Ia and Ib interneurons. Ia interneurons receive input from muscle spindles and send inhibitory signals to MNs innervating antagonist muscles \citep{pierrot2012circuitry, jankowska1972morphology}. Ib interneurons contribute to the inhibition of synergistic MNs and activation of antagonist muscles \citep{Eccles1964, Cote2018}. Additionally, complex multi-layered networks, including group II interneurons, control body posture during locomotion based on inputs received from dedicated supra-spinal structures~\citep{Drew2004, Sinha2020}.  

The supra-spinal structures involved in the adaptation and refinement of gait patterns are the brainstem, cerebellum and motor cortex \citep{pierrot2012circuitry}. The brainstem contributes to the modulation of MN excitability \citep{MacDonell2015} and regulation of gait patterns \citep{Whelan1996}, as indicted by animal studies involving electrical stimulation. Accordingly, cats presenting spinal cord injury where only able to modulate their locomotive patterns when the level of the spinal lesion did not disrupt the connection with the brainstem \citep{Whelan1996}. Moreover, the front part of the brainstrem produces dopamine, a neurotransmitter, which partly modulates the speed and direction of locomotion in treadmill-running animals \citep{Freed1985, Hulse1979}.

The cerebellum has been associated to locomotion adaptations, as suggested by experiments on cats where long-term depression of the cerebellum activity resulted in the lost of locomotor adaptive capability \citep{yanagihara1996nitric}. Additionally, the cerebellum was suggested to be responsible for the adaptations observed in the slower leg of human subjects  during split-belt treadmill studies \citep{Reisman2007}. 

The cortical motoneurons, located on the primary motor cortex of the brain \citep{Rathelot2009}, also contribute to the control of locomotion \citep{Barthelemy2010}. Experiments on cats have demonstrated that cortical lesion leads to reproducible and long-lasting deficits adapting to changing environments during gait, suggesting the brain’s involvement in voluntary gait modification \citep{Drew2002}. Single-neuron recordings from non-human primates have suggested that the motor cortex is involved in the termination of the stance phase and initiation of the swing phase \citep{Armstrong1985}. The activation patterns of cortical neurons have been shown to covary with locomotion kinematics \citep{zandvoort2022cortical}. Also, experiments on cats showed increased activation of the motor cortex when precise hindlimb control was necessary to step over obstacles during locomotion \citep{Drew1988}. Additionally, it has been suggested that the motor cortex is involved in the modulation of interneuronal circuits processing sensory information \citep{Pearson2000} and reflex responses \citep{Nielsen2016}. This brain-controlled modulation is demonstrated by experiments involving trans-cortical reflex loops in leg muscles \citep{Christensen2000} and attenuation of stretch reflex to prevent destabilizing effects \citep{Shumway-Cook2007}.

\subsection{Sensory Feedback}
\label{sec:SensoryFeedback} 

Muscle spindles and Golgi tendon organs formalize the proprioceptive feedback loop of human motion control. Muscle spindles are associated with stretch-related characteristics such as length, velocity \citep{Matthews1963, Matthews1969}. Golgi tendon organs encode force-related variables in active muscles \citep{Jami1985}. Recent studies have showed that muscle spindles are also directly related to changes in force in passive (relaxed) muscles \citep{Blum2017, SimhaTing2024}. Via afferent pathways \citep{Gervasio2017, Yavuz2018, Marque2001}, this information is transferred to different layers and functions of motor neuron pools. Stretch-related responses reach the pool of motor neurons through two main circuits arising from the muscle spindles: a direct, monosynaptic connection (via Ia afferent fibers) \citep{jankowska1972morphology}, and a multi-synaptic connection mediated by inter-neurons (via II afferent fibers) \citep{riddell2000interneurones} (Fig. \ref{Fig1_updated}). Force-related responses, on the other hand, reach the motor neuron pool indirectly from the Golgi tendon organs (via Ib afferent fibers) through spinal inter-neurons \citep{jami1992golgi} and long-loop connections that extend towards the brain and back \citep{jankowska1992interneuronal}.

The sensitivity of the muscle spindles is regulated by the fusimotor system \citep{Ellaway2002}. The static and dynamic components of the muscle spindle’s sensitivity are modulated via gamma-motoneurons ($\gamma$-MNs) receiving input from spinal and supraspinal structures \citep{Hulliger1989}. Although $\gamma$-MNs have been reported to co-activate with $\alpha$-MNs \citep{granit1955receptors}, particularly in the case of locomotion \citep{sjostrom1976muscle, bessou1989discharge}. Experiments on decerebrated cats suggest a more flexible control of $\gamma$-MNs \citep{ellaway2002role}, where the properties of the muscle spindle adapt to the demands of skeletal muscles during locomotion.

\subsection{Muscle-Tendon units}
The muscular system is highly redundant. There are more than 600 muscles that span a much lesser number of biological joints \citep{lacquaniti2012patterned}. The muscles are both redundant, with multiple muscles acting on the same joint, as well as multi-articular, i.e., crossing multiple joints \citep{hogan1985impedance}. Eventually, the neural control results in muscle forces, that results in mechanical actuation of the skeleton. Muscles have high power-to-weight ratios, high efficiency, and are capable of generating forces of thousands of newtons in less than a second \citep{uchida2021biomechanics, edwards1981human}. Muscles and tendons function as a mechanical unit around biological joints. Tendons are passive and elastic elements that play a crucial role in storing and returning elastic potential energy, thereby improving efficiency in quasi-periodic tasks such as walking. 

Muscles follow a hierarchical structure across spatial scales, with the sarcomere as a functional force-generating unit. Sarcomeres generate force in the pulling direction, i.e., shortening direction, via interactions of three main filaments: actin, myosin, and titin \citep{herzog2017skeletal, eisenberg1980relation}. In addition, the force-generating capacity of muscles depends non-linearly on their operating lengths, contractile velocities, and contraction history-dependent properties, such as fatigue  \citep{hawkins1997muscle}, residual force enhancement and residual force depression \citep{hahn2023history}. Muscles receive active commands, i.e., action potentials (APs), from spinal motor neurons \citep{heckman2012motor}. With the accumulation and release of $Ca^{2+}$, muscle force twitches are generated  \citep{dulhunty2006excitation}. As muscles do not completely reset to their relaxed state with increase in firing rate between successive APs, the resulting muscle forces are a smoothed outcome of the APs \citep{partridge1965cyclic}. Muscles undergo structural changes in response to aging or sarcopenia, or to exercise, thereby further altering their contractile properties across time \citep{evans1995sarcopenia, davies1986mechanical}. 

\section{Neuromechanical Models of Movement} \label{sec:2}
Locomotion requires a harmonic accord of tonus of multiple musculotendinous systems controlling the torque at multiple joints. This harmony is orchestrated by dexterous cooperation between neural information and musculoskeletal mechanics that mutually adapt to each other. 

Although movement planning starts in the brain, animal studies show that much of gait control is automatic \citep{cohen1980neuronal, delvolve1999fictive, soffe1982tonic}, involving local circuits (CPGs) dedicated to cyclic movements like walking and running. When impaired by neurodegenerative diseases, spinal injury, stroke, or amputation, the body loses crucial sensory feedback for gait control. Lower-limb prostheses struggle with postural instability, spinal deformity \citep{Hendershot2013}, and chronic pain due to this loss. A solution is to convey sensory information from the prosthesis directly to the user's nervous system, reintegrating it for balance control and embodiment sensation. This can be achieved with novel neural prostheses \citep{Koelewijn2021}.

The stretch reflex mechanism (muscle spindle and GTO) and the fusimotor mechanism couple the mechanical system with neural information. Neural drive and motor output adjust acutely or chronically due to factors like aging \citep{Enoka2003}, fatigue \citep{Gandevia1996}, pain \citep{Yavuz2015}, neuromuscular degeneration \citep{Schmied1999}, or spinal cord injury \citep{Gogeascoechea2020}. The system also adapts muscle tone and neural control based on tasks \citep{Laine2014}. Wearable robots must consider these adjustments and adapt their assistance based on feedback from the residual NMSK system.

Thus, to push the boundaries of WRs, we require models of the NMSK system that are personalized as well as easily implementable on the WR \citep{Mahdian2023}. Here, we summarize the state-of-the-art approaches to model the neuromechanics of locomotion, which encompasses both gait and balance. We address model-based techniques that target the neural and muscle systems and also data-based approaches. 

\subsection{Neural Models}
Mathematical models of the motor units, CPGs, spinal reflexes, and synergies are presented here.

\subsubsection{Motor Units}
Neural input is amplified via muscles to muscle activity during a movement. EMG signals are thus a mixture of electrical activity from multiple motor units. As motor units are fundamental elements for movement, decoding their behavior is essential to create more precise and physiologically correct musculoskeletal models. Recent advancements in high-density surface EMG decomposition \citep{holobar2007multichannel,clarke2020deep, negro2016multi} allow interfacing with motor units in vivo.  In turn, this enables the generation of better activation dynamics to estimate muscle force \citep{CailletModenese} and joint moments \citep{sartori2017vivo}. However, current decomposition techniques are mostly limited to isometric contractions.      

Additionally, recent studies \citep{Ornelas-Kobayashi, CailletModenese} showed that computational MN models driven by non-invasively estimated common input currents (i.e., linear transformation of the neural drive \citep{Farina2014}) can be calibrated to reproduce experimental spike trains and estimate the output of complete MN pools. These approaches enable the estimation of neuro-anatomical features in a person-specific way (e.g., MN sizes and ionic channel dynamics) \citep{Ornelas-Kobayashi} and, due to an increased amount of quantifiable MNs with respect to decomposition alone (i.e., from few tens \citep{MerlettiAndHolobar} to hundreds \citep{Ornelas-Kobayashi, CailletModenese}), have the potential of improving prediction accuracy of muscle activation dynamics.

These modeling approaches could help us better decode the neural intention during movement. 

\subsubsection{Central Pattern Generators}
Human bipedal locomotion could be generalized as a cyclical movement using CPGs, which is a simplistic numerical model. CPGs approximate the neural muscle control that generates the basic rhythm and patterns of the spinal motoneuron activations during walking \citep{lacquaniti2012patterned}. Rhythmic activities in  lampreys \citep{cohen1980neuronal}, salamanders \citep{delvolve1999fictive} or frog embryos \citep{soffe1982tonic} are generated by the nervous system with no inputs from the sensory feedback, and can be useful for bio-inspired robotics \citep{Ramdya2023}. The presence of spinal CPGs in humans is supported by studies that show that new-borns can walk with some support despite the absence of a mature connection between the cortical and spinal region \citep{dominici2011locomotor,grillner1985neural,martin2005corticospinal}.

There are different approaches to model CPGs. Gentaro Taga \citep{taga1994emergence} represented CPGs as a network of oscillators coupled with the environment. \citet{Dzeladini2014} demonstrated via neural modeling that CPGs can be beneficial to modulate speed/step length during the gait and can be seen as feedback predictors. Based on physiological studies \citep{burke2001patterns,lafreniere2005deletions,rybak2006modelling}, CPGs were modeled as a two-layered model characterized by a rhythm generator, that can modify the basic rhythm of the gait by phase shift as a response to sensory feedback, and a pattern formation network \citep{aoi2016neuromusculoskeletal,ivanenko2005coordination,ijspeert2008central}. Other approaches to emulate the biological role of CPGs are simulated by adaptive oscillators (AOs) \citep{aoi2010evaluating, ruiz2017experimental}. An example of CPGs are shown in Figure \ref{fig:NeuralModels}.

CPG models have the advantage of reducing the dimensionality of the control problem and providing a quick response to walking condition changes such as speed or inclination transitions. However, the extent of modulation of muscle activations from sensory feedback during adaptive walking is still a topic of debate. One study \citep{tamura2020contribution} investigated the role of the phase resetting algorithm combined with a synergy model to address changes in walking conditions. Their results support the hypothesis that timing changes could play an important role during transitions such as changes in speed, stepping over an obstacle or changes in elevation of the ground. Moreover, recent neurological studies \citep{zandvoort2022cortical} show the correlation between the cortical region and pattern of activation that are developed in adult age for independent walking.
Indeed, if we assume the gait is a result of the interplay between the central nervous system, the musculoskeletal system, and the environment, a CPG model only covers part of that interaction. Preliminary investigations included both feedforward and feedback signals to drive a musculoskeletal model, their simulation results showed that while \textit{reflexes} are fundamental to address perturbed walking, CPGs regulate the periodicity of the motion, therefore it seems that both are required to have a complete model of human motor control\citep{Russo2023}.

\subsubsection{Spinal Reflexes}

Spinal reflexes are partly responsible for muscle activations in voluntary locomotion, generating automatic responses from muscle stretch, velocity, and force among other inputs. At the muscle spindle and fiber level, mathematical models have been developed to capture those detailed responses from experimental recordings \citep{Maltenfort2003, Mileusnic2006, Harding2018}. Reflexive mathematical models for locomotion, in turn, aim at describing the general muscle activations that control the behavior of lower limbs during locomotion \citep{Geyer2010}\citep{Song2015}. The set of individual reflexes that compose the overall reflex model differ in their internal structure and in the stimuli that inform them. Structures vary in how inputs are connected to outputs and by how outputs are generated (linear or non-linear relationships), whereas inputs were originally described as body and muscle states and leg loading and later expanded to include additional information such as the center of mass states. A generic formulation of reflexes is provided in Appendix A.

\begin{figure*}[t!]
   \centering
    \includegraphics[width=\textwidth]{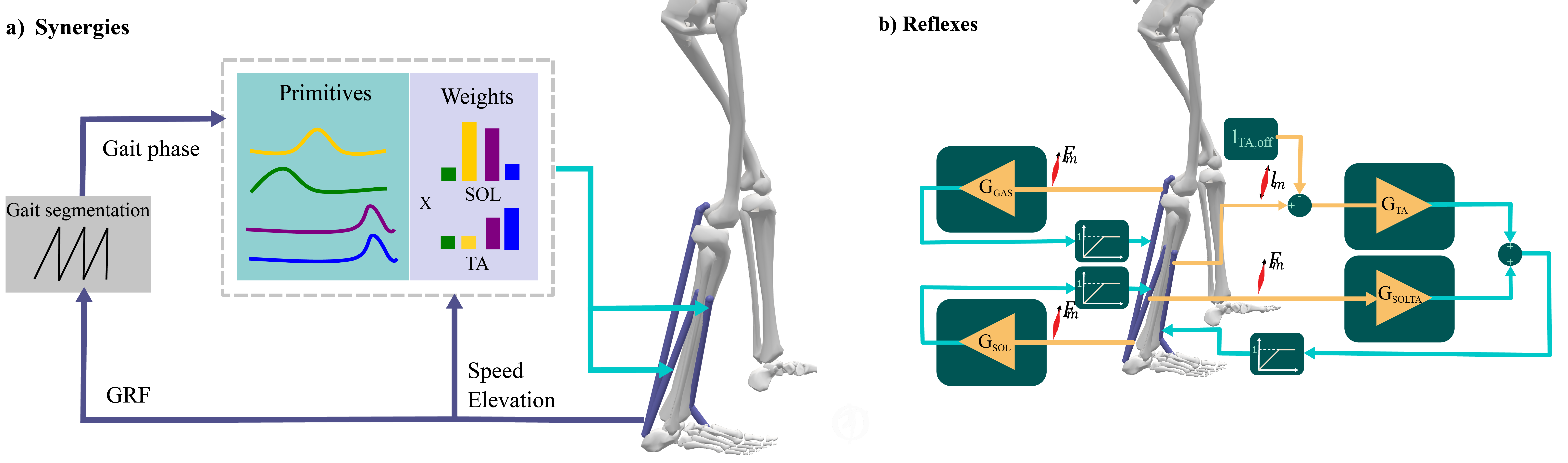}
  \caption{Modeling a) synergies and b) reflex \citep{Geyer2010} of select muscles of the lower leg. Synergies model contains a primitive which receives the gait phase information and a weight block which receives sensory input to adjust the weights. In the reflex model, gastrocnemius and soleus muscle activations during the stance phase result from a positive force feedback loop derived from the respective muscles. Tibialis anterior activation results from offset length feedback and suppressive force feedback from the soleus muscle. Feedback loops are saturated between 0 and 1.}
  \label{fig:NeuralModels}
\end{figure*}

The most common feedback types in a reflex model are muscle force feedback and muscle stretch or fiber length feedback \citep{Geyer2010} as a model of reflexes generated as responses to stimuli from muscle spindles and GTOs. 
Feedback could be provided by either the muscle that generates the activation or other muscles that influence this muscle. Neural delays are also incorporated into the model. Other studies also included joint angle and center of mass states to improve the match with experimental data \citep{Geyer2010, Keemink2021}. Fig. \ref{fig:NeuralModels} provides an illustration of the reflex model during the stance phase for select muscles of the lower leg.

Early descriptions of human locomotion, comprised of a reflexive and central pattern generation \citep{Ogihara2001}, were successfully reduced to a model containing reflexes exclusively, while generating stable human-legged locomotion \citep{Geyer2010}, implying that supra-spinal input plays a minor role in the generation of cyclic walking behavior. Muscle fiber lengths, muscle forces, and joint angles are proprioceptive inputs to the model, whereas leg load is the only external input. The usage of a higher-level input layer is explored in several studies \citep{Desai2013, Dzeladini2014, Ferreira2015, Greiner2018, Haeufle2018, Harding2018, Ramadan2022, Sar2020, Schreff2022, Suzuki2018, VanDerNoot2015, VanderNoot2018, ramadan2022neuromuscular, di2023investigating}. Higher-level inputs can, for instance, be used to assure leg clearance from the ground or to influence step length. The model in \citep{Geyer2010}, when evaluated in a musculoskeletal simulation, generates human walking kinematics that tolerates a small number of ground perturbations, including unexpected slopes. Joint angles during locomotion bear striking resemblance if compared with human subjects in the same conditions. Muscle activations however are smoother in appearance and show some significant disparities in both amplitude and timing. 

Later reflex models were proposed to expand their initial capabilities into generating realistic locomotion movement in 3D space \citep{Song2015} and during pelvis perturbations. In it, additional muscles and two new feedback types (fiber velocity and co-stimulation) are proposed. Fiber velocity inputs may be used to dampen the movement of the knee segment by modulating the activation of flexor muscles. Co-stimulation, on the other hand, is used for antagonist muscle suppression, for positive feedback on the muscle itself, or for group activation of muscles with similar functions. Additionally, supraspinal inputs were provided in the form of target leg clearance from the ground and foot placement targets in the form of the angle formed by the ground plane and the segment that connects the desired foot position and the hip segment of the model. Finally, instead of two distinct phases: stance and swing, the swing phase were subdivided into 4. The proposed model improved in its capacity to maintain stability after moderate pushes at the pelvis, to withstand ground obstacles, and to go upstairs, to turn, and to run. The sensitivity of those models to sensory noise \citep{Brug2017} and speed \citep{Song2012, Song2015} were also investigated.

Balance is also hypothesized to be partly controlled by reflexive pathways \citep{Afschrift2021, Geyer2017, Keemink2021, Koelewijn2020}. By adding center of mass velocity as feedback to the activation of muscles that drive the ankle joint, studies have demonstrated that the activation of such muscles and the resulting ankle torques are more in line with its human counterparts under the same conditions \citep{Afschrift2021, Keemink2021, Afschrift2022}. This indicates that reflexes are significant in generating muscle activations as a response to external perturbations.

\subsubsection{Synergies}
The human body has many more muscles than its degrees of freedom \citep{lacquaniti2012patterned}. Groups of muscles contribute to the execution of the same action by cooperation and synchronization. During locomotion, the nervous system organizes motor control into a small number of coordinative modules (i.e. a group of muscles) or muscle synergies to reduce the complexity of the control problem \citep{bizzi2013neural,lacquaniti2012patterned}. For example, the muscle activation signals of leg muscles during human adult locomotion can be reconstructed using four to five principal components \citep{lacquaniti2012patterned}. The components can further be grouped into agonists or antagonists which determine their activation pattern during the movement.

There are several methods to extract the synergy information from experimental studies. For example, muscle force information measured experimentally can be used to identify muscle synergies for a given set of tasks \citep{iansek2013rehabilitation, Ting2012}. \citet{meyer2016muscle} proposed an optimization method to match the synergy calculated from statistical analysis of joint moments, joint kinematics, and ground reaction forces which could be useful for movement prediction and control\citep{Michaud2020}. Recently, \citet{scano2022mixed} proposed a new method called mixed matrix factorization that can limit the constraints to have only non-negative factors and include the relation between kinematic events and muscle activity. Another study proposed deriving primitives and corresponding weights from the joint torques, which allows relating the synergies to the biomechanical outputs  \citep{gopalakrishnan2014novel}. A generic formulation is provided in Appendix B. 

Statistical methods of synergy extraction select the number of synergies based on variance (VAF). This may not be reliable because of either the statistical methodology used or the quality and quantity of the input data. Therefore, evaluating the functional outcome of the extracted synergy becomes challenging \citep{Turpin2021, Ballarini2021}. Moreover, the number of synergies is dependent on the number of skeletal muscles recorded which results in discrepancies across studies \citep{Moiseev2022, Funato2023}. The theory of muscle primitives is based on the hypothesis that muscle excitations include a spatial and a time contribution: the latter is given by the weights associated with each muscle for the specific pattern  \citep{lacquaniti2012patterned}. Therefore, muscle primitives are modeled as a set of impulsive curves \citep{neptune2009modular,allen2012three,mcgowan2010modular}, rectangular pulses \citep{aoi2016neuromusculoskeletal,garate2016walking,aoi2010evaluating}, or Gaussian curves \citep{sartori2013musculoskeletal}. \citet{sartori2013musculoskeletal} advanced these models by including muscle primitives that were shaped to match experimental ones. Furthermore, \citet{gonzalez2015predictive} analyzed the gait movement at different speeds and inclination conditions using non-negative matrix factorization from recorded EMG signals \citep{rabbi2020non}. They demonstrated that the activity of leg muscles can be reconstructed as a linear combination of four basic patterns, whose peak of activation depends on the phase of the gait cycle. The average shape of each pattern once time normalized to the stride duration, is little affected by changes in speed, direction, loading, and task \citep{gonzalez2015predictive}. Further, they used this model combined with a musculoskeletal model to predict experimental joint torques with a static optimization method to adjust the weight of the primitives \citep{sartori2017predictive}. Others demonstrated that by changing a few parameters in the model it is possible to reproduce walking and running patterns at different speeds \citep{aoi2019neuromusculoskeletal, damonte2023synergy}. 

Synergy models have also been utilized to predict the muscle excitation of deeply located muscles and to overcome the limitation of signal recording from surface EMGs \citep{ao2020evaluation,ao2022emg,sartori2014hybrid}.

\subsection{Muscle Models}

Modeling the musculoskeletal system has been accelerated with advances in multi-dynamic system models. This has also led to several computer simulation models. Here, we briefly describe the state-of-the-art modeling for the musculoskeletal system. 

\subsubsection{In-silico models}
\emph{In-silico} representations of muscle dynamics can be modelled either mechanistically or phenomenologically \citep{rohrle2019multiscale}. Mechanistic or biophysical models accurately capture the anatomical structures and physiological processes underlying muscle force generation. These models require detailed \emph{a-priori} knowledge of muscle physiology and comprise a large set of parameters and differential equations. Therefore, despite their ability to investigate cause-effect relationships between muscle activation and contraction dynamics, these models are computationally expensive and, consequently, are not suitable for control purposes. One example is the Huxley model, that simulates the contraction of a sarcomere based on the Huxley's sliding filament theory \citep{huxley1957muscle}. It considers the interactions between myosin and actin filaments, as well as the activation of calcium ions, to predict the force and length changes in a sarcomere. Phenomenological models are rather based on empirical input-output relationships. They can capture the muscle behaviour within the conditions that have been used to obtain experimental data, and they comprise a smaller set of parameters and are usually computationally efficient, making them a feasible choice to simulate muscle contractions for control purposes. However, phenomenological models can provide limited answers to understand underlying physiology. The most common phenomenological model is the Hill-type muscle model \citep{hill1938heat}. It has been widely used in various analyses and simulation studies \citep{robertson2013research, mcmahon1984muscles, scott2004optimal, samozino2016simple, Buchanan2004}. It captures the major nonlinear properties of biological muscles, such as activation dynamics (low-pass filter effect), nonlinear force-length/velocity relationships, maximum voluntary contraction forces, and the nonlinear tendon force-strain relationship. 

\subsubsection{Simulation models}

Building mathematical models of high-dimensional multi-dynamic systems is challenging. Therefore, current models address specific parts of the function of the human neuromusculoskeletal system. There are several standard human body models \citep{Seth2018, caggiano2022myosuite} in different simulation platforms, such as OpenSim\footnote{https://simtk.org/projects/opensim/}, AnyBody\footnote{www.anybodytech.com}, and MuJoCo\footnote{https://mujoco.org/}. These platform provide high fidelity information on musculoskeletal system and contact with external environments. In 2D gait studies, the seven linkages and nine degrees of freedom (DoF)  mathematical model is most commonly used \citep{Ackermann2010, Geyer2010} to represent body dynamics. In 3D studies however, models with different levels of details are used to address different movement conditions. For instance, the Rajagopal full-body model (20 segments and 40 DoF) \citep{Rajagopal2016} helps to understand the full-body effects on fast locomotion types, such as running and other arm-involved movements \citep{renganathan2022effect, mohr2021sex, mahadas2019biomechanics, song2021deep}. Alternatively, a model without arms \citep{au2013gait} is typically used in slow movement, such as walking \citep{de2021perspective, weng2022adaptive, de2016evaluation, de2008kalman}. Other modeling platforms such as SCONE\footnote{https://scone.software/} \citep{Geijtenbeek2019} or HyFyDy\footnote{https://hyfydy.com/} target performance and speed allowing model simulations based on high-level objectives such as energy or walking speed. 

\subsubsection{EMG-informed models}
EMG recording has been around for more than a century \citep{Reymond1843}. Surface EMG sensors are cheap and easy to use but can only record muscle close to the skin and is prone to noise and muscle cross-talk. EMG data is generally processed by using a standard pipeline of filtering and extracting the envelope, which is subsequently normalized against maximal voluntary contraction to obtain the normalized envelopes \citep{Winter2009}. This information can be used to estimate the activation dynamics of the respective muscles, which in turn can be used to extract muscle force  information. Direct recording removes the need for assumptions on muscle coordination (synergies) or feedback gain (reflex) allowing volitional and continuous accessing of neural control outputs (commands).

EMG-informed models have been applied to multiple joints and muscles to cover complex movements \citep{Lloyd2003, Buchanan2004, Sartori2012}. Furthermore, experimental EMG paired with inverse kinematics and dynamics allows for personalization of the musculoskeletal model  which is not possible with a surrogate of the neural systems \citep{Pizzolato2015}. Real-time implementation of these models were developed quite recently, which offers a new opportunity for controlling wearable devices \citep{Durandau2018, pizzolato2017biofeedback}.
These studies show that EMG-driven models can be executed faster than the electromechanical delay of the muscle, have accurate joint torque estimations, and further extrapolation capability on tasks and DoFs not used during the calibration process \citep{Durandau2018, Durandau2022, Moya-Esteban2023}. The open-source nature of these toolboxes, such the Calibrated EMG-Informed NMS Modelling Toolbox (CEINMS)\footnote{https://simtk.org/projects/ceinms} can reduce the barrier in implementing such approaches across research labs  \citep{Pizzolato2015}.\\

\subsubsection{Muscle-tendon kinematics based models}
Measuring muscle or tendon kinematics provides a direct measure of the internal states of the muscle without requiring models that estimate them. This information can be obtained using ultrasound or tensiometry. However, these techniques are complex to use, and are sensible to sensor noise and movement and require more complex processing. New machine learning (ML) algorithms have simplified tracking of muscle fiber length \citep{cronin2020fully} and have been implemented for real-time applications \citep{rosa2021machine}. This information can then be used within a muscle model to extract muscle force and joint torque for control of WRs \citep{nuckols2021individualization}. Shear tensiometry is a novel approach that allows allows measuring forces of superficial tendons \citep{martin2018gauging}. This technique was later used to quantify change during exoskeleton assisted walking and the reduction of tendon force during assistance \citep{schmitz2022modulation}.
 
\subsection{Data-driven Models}
As the generation of muscle force and its translation to movement is a highly non-linear system, the mapping has been performed using ML approaches. \citet{geijtenbeek2013flexible} explored the effectiveness of a ML based controller for bipedal locomotion. Simulation results showed that ANN-based controllers were able to generate suitable muscle activations to drive musculoskeletal models to achieve stable locomotion gaits. In addition, ANN-based controllers have been used to generate pathological gait patterns, amputee locomotion gaits with passive prostheses, and gait patterns with orthopedic surgeries \citep{lee2019scalable}.
For instance, spiking neuron pools informed artificial neural networks (ANNs) have been used to model the strong nonlinear and multi-functional properties of the central nervous system \citep{abbas1995neural, iyengar2019curated, iyengar2021novel, lan1994neural,  sreenivasa2013modeling, zhong2021bioinspired}. Sreenivasa et. al. identified a realistic spiking neural network that included dynamics of spiking motor neurons and muscle fibre bundles (motor units), spindle afferents, intra-spinal connections and skeletal mechanics, for the stretching reflex control. Simulation results showed that the identified model can reproduce the stretching reflex movements \citep{sreenivasa2013modeling}.

Neuronal model platforms NEURON and the musculoskeletal co-simulation platform NEUROiD have been developed and been applied to lower leg functional tasks \citep{hines1997neuron, iyengar2019curated}. Current neuronal control platforms focus on the spinal neuron connections with the afferent feedbacks, such as from the Ia, II, and Ib fibers. Thus, deep learning was integrated with these neuronal control platforms to model higher level intention from the brain \citep{iyengar2021novel}. \citet{Vargas2024} recently employed data-driven models to learn an internal representation of how proprioception is coded at the neural level. These models were informed by musculoskeletal modeling which allowed testing different initial hypotheses of the representation using simulated data.

\section{Neuromechanical Control of Wearable Robotics}
\begin{figure*}[t!]
  \centering
    \includegraphics[width=0.8\textwidth]{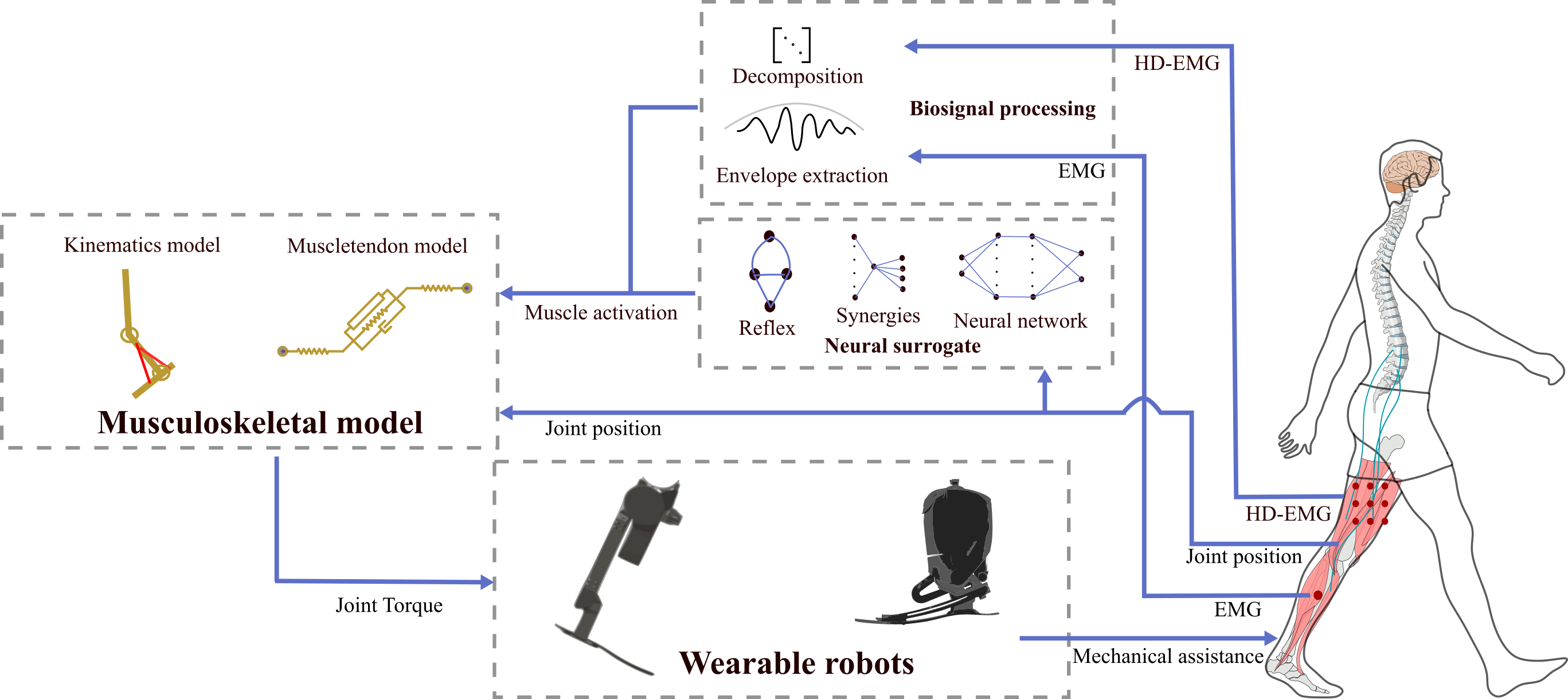}
  \caption{Possible approaches to integrate neural models and musculoskeletal models to control wearable assitive robots.}
  \label{fig:WR}
\end{figure*}

In this section, we summarize state-of-the-art controllers that utilize the models we have summarized in Section \ref{sec:2}. A graphical overview is provided (Fig.\ref{fig:WR}). We describe controllers that interface with the neural system, use musculoskeletal models, integrate human-in-the-loop, or use data-driven approaches. 

\subsection{Neural controllers}
Current control schemes for wearable robotics commonly target metabolic output \citep{zhang2017human}, muscle kinematics \citep{nuckols2021individualization} and tendon force \citep{schmitz2022modulation}, and disregard neural control. Although a few studies use an indirect estimation of the neural control, such as EMG \citep{sawicki2009pneumatically, Durandau2022}, they offer poor insight into the patient adaptation and change at the neural side due to the assistance or rehabilitation caused by the wearable devices. This insight could have potential advantages in interfacing with people who suffer from a neural injury such as stroke or spinal cord injury (SCI).

A few studies have implemented neural interfacing and shown interesting results, including use of a reflexive controller to assist SCI patients \citep{Wu2017} or prostheses \citep{eilenberg2010control}. Using relatively few inputs (joint angles and gait phases), a healthy torque pattern can be decoded when used in collaboration with a NMSK model, which can be used to assist the user. There are a few limitations of this method, mainly the inability to voluntarily start and stop, need for optimization of the reflex’s parameters to gait only, and reliance on healthy neural control. Compare to reflex-based controllers, synergy-based controllers \citep{ruiz2017experimental} offer similar advantages and suffer from similar shortcomings. The main difference is that synergy requires less input (gait phase only). Although, this cannot account for external perturbations, which are represented as feed-forward control. On the contrary, reflex-based controllers could be suitable for perturbation \citep{keemink2021whole} as they represent a feedback control. 
Neural-based controllers are currently being investigated for their simplicity and ability to generate stable gaits. However, as these gaits represent surrogates of healthy neural control, it must be tailored for non-healthy gaits. An alternative would be to combine the feed-forward aspect of synergy with the feedback aspect of the reflex-based control to create more robust and personalizable controllers \citep{Russo2023_reflexes}.

\subsection{Muscle model based controllers}
WRs do not interact directly with the user’s nervous system but rather with human tissues (bones, muscles, tendons and skins) via mechanical interaction. Therefore, WRs must translate neural information into resulting mechanical control. EMG signals can be used as a surrogate of the neural system and can be used as a direct control signal. One such study utilized this approach for a pneumatically powered orthosis \citep{sawicki2009pneumatically}. The provided mechanical force was able to assist during plantar flexion and showed the possibility of reducing EMG activity of the user’s soleus. A drawback of this approach was the need to find optimal gain parameters which could vary between sessions, tasks, and users. Another limitation was the difficulty of recording EMG signals with low signal-to-noise ratio, that are needed for the control scheme. To solve for these limitations, measured muscle activity could be integrated with a musculoskeletal system of the joint of interest \citep{Durandau2022}. This can provide the biological joint moments which can then be used by the assistive devices as a control signal. Such an approach has been shown to reduce muscle activity during different types of gait \citep{Durandau2022}. Such a model are typically driven with EMG envelopes which offers only the amplitude information about the muscle activity. Novel techniques to extract the neural characteristics from HD-EMG could offer a better interfacing with the devices and must be explored for locomotion and balance \citep{Caillet2024}.

\subsection{Human-in-the-loop based controllers}
The central idea behind bio-inspired neuromuscular control approaches is to obtain assistive torques that are, to a large extent, human-like. This has been shown to be effective in reducing muscle activation during walking \citep{Durandau2022} and during recovery of balance during perturbed standing \citep{Emmens2018} and walking \citep{Afschrift2022}. 
Several groups applied human-in-the-loop (HIL) optimization to find time-based torque profiles to reduce the metabolic costs of walking. A review \citep{Sawicki2020} of these studies shows how various autonomous wearable hip and ankle exosuits resulted in metabolic cost reductions of up to 20\%. The timing of the estimated assistive torque profiles appeared to be critical and different from physiologically measured torque profiles. Larger assistive torque did not necessarily result in a larger reduction of metabolic costs. \citet{Beck2023} found that standing balance can be improved when an ankle exoskeleton provides support faster than physiological responses. Muscle dynamics and neural reflex loops are characterized by low-pass dynamics and significant time delays in the ascending and descending pathways. Thus, approaches that don’t attempt to mimic physiological neuromechanics might result in faster and more powerful assistance, thereby overcoming inherent biological limitations resulting in super human performance. For this reason, it might be interesting to investigate the effect of human-machine interfaces that still use the neuromechanical modelling framework, but with supra-human neuromechanics such as faster reflexes.

\subsection{Data-driven approaches for control}
ML approaches to control wearable robots are quite common \citep{Liu2024}. However, these models might not be generalizable for tasks beyond the training dataset. Some studies have rather utilized ML approaches to optimize parameters for design of controllers for the WRs. This includes the use of reinforcement learning to learn hyperparameters of the controller for personalization \citep{Berman2023}, or utilizing human-directed reinforcement learning to improve user experience with the wearable robot \citep{Lee2023, molinaro2024estimating}, or reinforcement learning methods to design robust controllers  \citep{luo2024experiment}. A recent seminal work showed the feasibility in transferring control policies developed using ML approaches applied to simulated data towards implementation in an exoskeleton \citep{Luo2024}. The learning algorithm implemented model based musculoskeletal and exoskeleton models as well as their interactions. This shows the promise of using data-driven approaches for generating control policies applicable for real-world scenarios.

\section{Discussions}
In this paper, we present a comprehensive overview of methodologies for interfacing with the human nervous system in vivo and modeling the musculoskeletal system, particularly within the realm of WRs. We focused on three key aspects: the biological human NMSK system, mathematical models of neuromuscular controllers, and the current implementation of these controllers in wearable robotics. Our goal is to provide a common foundation for researchers, engineers, and clinicians interested in these topics. The challenge lies in developing NMSK models that are both computationally efficient for real-time human-machine interfaces and physiologically accurate for subject characterization.

In the first part on Neuromechanics of Movement, we detail the human NMSK system, explaining how the nervous system sends information to muscles and receives feedback to adjust control. We also describe how muscles actuate the skeletal system through tendons. We introduced the concept of muscles acting as biological amplifiers of neural output from the spinal cord. We also outlined techniques for modeling spinal neuronal networks controlling musculoskeletal force generation, including biophysical models of neurons forming spiking neuronal networks and phenomenological models of spinal feedforward and feedback mechanisms. 

In the second part on Neuromechanical Models of Movement, we explore various methods for modeling neuromuscular control. High-density EMG decomposition and modeling were highlighted as methods to decode synaptic input projected to motor neuron pools, facilitating the creation of comprehensive in silico motor neuron pools tailored to individual subjects. We examine established neural control models of walking, including central pattern generators and feedback-based pattern generators inspired by reflexes. We also discuss methods for directly accessing human neural control using sensors, eliminating the need for assumptions. Additionally, we delve into novel neural control solutions using spiking neural networks and deep neural networks. 

In the final section on Neuromechanical Control for Wearable Robotics, we highlight the importance of integrating neuromusculoskeletal knowledge into robotic controllers, addressing challenges such as metabolic reduction and personalized assistance for rehabilitation. We demonstrated how neuro-muscular features derived from data-driven musculoskeletal models could be integrated with assistive robots for in vivo interfacing with the patient's neuromuscular system. This approach provides insights into underlying neuro-mechanical relationships, their variations with pathology, or changing muscle properties, and their potential generalization to novel conditions. 

WRs have traditionally been developed by mechatronic and control researchers applying control theory solutions that are effective in robotic systems. These solutions often overlook the human biological system, which can result in controllers that are either ineffective or counterproductive. The field of biomechanics has contributed to creating more efficient controllers by considering the human body’s mechanics, such as muscle and tendon actions during walking. This has led to controllers that significantly reduce metabolic costs and provide efficient assistance to healthy individuals. However, these metabolic reductions do not necessarily benefit individuals with neural or musculoskeletal injuries. Understanding the NMSK system in robotic controllers is crucial for developing effective rehabilitation solutions.

Many of the presented works are based on studies of healthy individuals or models. Further research on how diseases disrupt these models is essential for developing wearable robotics that provide targeted assistance for rehabilitation. Additionally, the personalization of models is critical, as each individual and injury is unique, highlighting the need for establishing realistic and individualized models. 

Future research should focus on adapting these models or sensor-based interfacing for WR controllers. This includes accounting for real-time high-density EMG decomposition algorithms and neural network simulations. Moreover, due to advances in image processing, vision could also be a viable inputs for WRs \citep{Kurbis2024, Gionfrida2024}. Another significant challenge is to incorporate changes in neural control over time to allow extended use of WRs during daily life. This might require the use of extensive human movement data, often termed "big data” \citep{Werling2023}.  The integration of wearable technology such as textile sensors, smartphones, watches, and orthotic devices, coupled with cloud-based data storage, promises to expand human movement data volumes to unprecedented scales.

Frameworks that integrate various modeling paradigms to process diverse data types, ranging from high-fidelity data acquired in laboratory settings to lower-fidelity data from wearable embedded sensors, will be crucial. Such integration will enhance the characterization of movement function and pathology with robust statistical inference and heightened predictive capacity. This evolution holds the promise of enabling increasingly accurate predictions, even in scenarios with limited or unavailable data. It may facilitate predicting the progression of neuromuscular or orthopedic diseases, anticipating responses to surgical interventions, or forecasting outcomes of physical training regimens, with significant implications for human-robot interfacing.

\section*{Acknowledgments}
The research was supported by the EU's RIA proposals SOPHIA (871237) and S.W.A.G (101120408), ERC Starting Grant INTERACT (803035), ERC Consolidator Grant ROBOREACTOR (101123866), and Marie Skłodowska-Curie Actions (MSCA) Innovative Training Networks (ITN) SimBionics (H2020-MSCA-ITN-2019-860850).

\appendix
\section{Spinal Reflexes}
Spinal reflexes are modelled as 

\begin{equation*}
A_{m,c}=\sum_{j=1}^J - [f(F_j (t-\delta_j),p_{1,j},\theta_{1,j} )+
g(l_j (t-\delta_j),p_{2,j},\theta_{2,j} )] + \sum_{i=1}^I - h_i(x_i (t-\delta_i))
\label{eq:reflexModel}
\end{equation*} \\

The activation ($A$) of a muscle ($m$) at a given phase ($c$) of the gait takes the generalized form described in equation \ref{eq:reflexModel}. 
The most common feedback types are muscle force feedback ($F$) and muscle stretch or fiber length feedback ($l$) \citep{Geyer2010} as a model of reflexes generated as responses to stimuli from muscle spindles and GTOs (Section \ref{sec:SensoryFeedback}). 
Feedback can come from the muscle itself ($j=m$) or other muscles ($j\neq m$), and it is governed by a gain ($p_1,p_2$) and a set of other parameters ($\theta_1,\theta_2$), commonly represented by a single offset on the input.

\section{Synergy models}
Primitives are modelled as Gaussian curves,
\begin{equation*}
    XP(t)=e^\frac{-(t-\mu)}{2\sigma^2}
\end{equation*} \\
Where t represent the gait cycle frame, $\mu$ is the time shift and $\sigma$ is the width of the Gaussian \citep{gonzalez2015predictive}. The activation timings of these patterns are strictly linked to specific kinematic events, such as foot contact and lift off.
Different excitations in time are the result of the sum of the four components weighted by some muscle-specific coefficient that is related to the specific walking condition(speed (v), elevation($\theta$).
\begin{equation*}
    MEP(v,\theta,t)=w(v,\theta)XP(t)
\end{equation*} \\
Where MEP is the Muscle Excitation Profile, v is the speed of the gait, $\theta$ the elevation of the ground, t the percentage of the gait.
Further they used this model combined with a musculoskeletal model in order to predict experimental joint torques with static optimization method to adjust the weight of the primitives \citep{sartori2017predictive}.
\citep{aoi2019neuromusculoskeletal} demonstrated that by changing a few parameters in the model it is possible to reproduce walking and running patterns at different speeds.\\

\newpage



%
\bibliographystyle{unsrtnat} 
\bibliography{References}

%




\end{document}